\pdfoutput=1
\documentclass[11pt]{article}

\usepackage[final]{acl}
\usepackage{array}
\usepackage{hyperref}
\usepackage{xfrac}
\usepackage{xurl}
\usepackage{times}
\usepackage{latexsym}
\newtheorem{definition}{Definition}
\usepackage[T1]{fontenc}
\usepackage[utf8]{inputenc}
\usepackage{microtype}
\usepackage{inconsolata}
\usepackage{subcaption}
\usepackage{xspace}
\usepackage{amsmath}
\usepackage{amssymb}
\usepackage{graphicx}
\usepackage{booktabs}
\usepackage{amsmath,array}
\newcolumntype{C}[1]{>{\centering\arraybackslash}m{#1}}
\newcolumntype{R}[1]{>{\raggedleft\arraybackslash}m{#1}}

\usepackage{tcolorbox}
\usepackage{inconsolata}
\usepackage{multirow}
\definecolor{c1}{cmyk}{0,0.6175,0.8848,0.1490} 
\definecolor{c2}{cmyk}{0.1127,0.6690,0,0.4431} 
\definecolor{c3}{cmyk}{0.3081,0,0.7209,0.3255} 
\definecolor{c4}{cmyk}{0.6765,0.2017,0,0.0667} 
\definecolor{c5}{cmyk}{0,0.8765,0.7099,0.3647} 
\definecolor{forestgreen}{HTML}{397727}

\newtcbox{\hlprimarytab}{on line, rounded corners, box align=base, colback=green!10,colframe=white,size=fbox,arc=3pt, before upper=\strut, top=-2pt, bottom=-4pt, left=-2pt, right=-2pt, boxrule=0pt}
\newtcbox{\hlsecondarytab}{on line, box align=base, colback=red!10,colframe=white,size=fbox,arc=3pt, before upper=\strut, top=-2pt, bottom=-4pt, left=-2pt, right=-2pt, boxrule=0pt}
\newtcbox{\hlterciarytab}{on line, box align=base, colback=gray!10,colframe=white,size=fbox,arc=3pt, before upper=\strut, top=-2pt, bottom=-4pt, left=-2pt, right=-2pt, boxrule=0pt}

\newtcbox{\hlgraytab}{on line, rounded corners, box align=base,colframe=white,size=fbox,arc=3pt, before upper=\strut, top=-2pt, bottom=-4pt, left=-2pt, right=-2pt, boxrule=0pt}

\newcommand{\dyes}{$D_{\checkmark}$\xspace}
\newcommand{\dno}{$D_{\times}$\xspace}

\title{NLP Systems That Can't Tell Use from Mention Censor Counterspeech, but Teaching the Distinction Helps}

\newcommand{\aspace}{\hspace{1em}}

\author{Kristina Gligorić \aspace
Myra Cheng \aspace
 Lucia Zheng \aspace
 Esin Durmus \aspace 
   Dan Jurafsky \\ Stanford University \\
   \texttt{gligoric@cs.stanford.edu} \\  
  \\}
  
\begin{document}
\maketitle

\begin{abstract}
{{\color{red!70!black} {Warning: content in this paper may be upsetting or offensive.}}}

%Distinguishing the use of language, which conveys meaning, from the mention of language,  which illustrates properties of language itself, is central to text-based inference.  The use-mention distinction is particularly essential to processing counterspeech that refutes problematic content,  as such speech often mentions harmful language (e.g., calling a vaccine dangerous is not the same as expressing disapproval of someone for calling vaccines dangerous.)  We investigate whether NLP models can distinguish use from mention, showing that even recent language models can fail at this task.  We then show that this failure propagates to two key downstream tasks, misinformation and hate speech detection, thus unintentionally censoring counterspeech.  We design and test prompting mitigations that teach the use-mention distinction and show that our interventions lead to a significant reduction in error.  Our work highlights the importance of the use-mention distinction for NLP and CSS and offers ways to address it.

The use of words to convey  speaker's intent is traditionally distinguished from the  `mention' of words for quoting what someone said, or pointing out properties of a word.  Here we show that computationally modeling this use-mention distinction is crucial for dealing with counterspeech online.  Counterspeech that refutes problematic content often mentions harmful language but is not harmful itself (e.g., calling a vaccine dangerous is not the same as expressing disapproval of someone for calling vaccines dangerous). We show that even recent language models fail  at distinguishing use from mention, and that this failure propagates to two key downstream tasks: misinformation and hate speech detection, resulting in censorship of counterspeech.  We introduce prompting mitigations that teach the use-mention distinction, and show they reduce these errors.  Our work highlights the importance of the use-mention distinction for NLP and CSS and offers ways to address it.
\end{abstract}

\section{Introduction}\label{sec:intro}
%par1: general background about the distinction
The \textbf{use-mention distinction} is the difference between using words ({\em Bananas have a peel}) and mentioning them (\textit{``Bananas'' has 7 letters}, or {\em Dan said ``Bananas''}).
The distinction has long been important in the philosophy of language~\cite{sperber_irony_1981,saka_quotation_1998}, where discussions date back to \citet{tarski1931concept} and \citet{quine1940use}.  

\begin{figure}[t]
    \centering
    \includegraphics[width=\columnwidth]{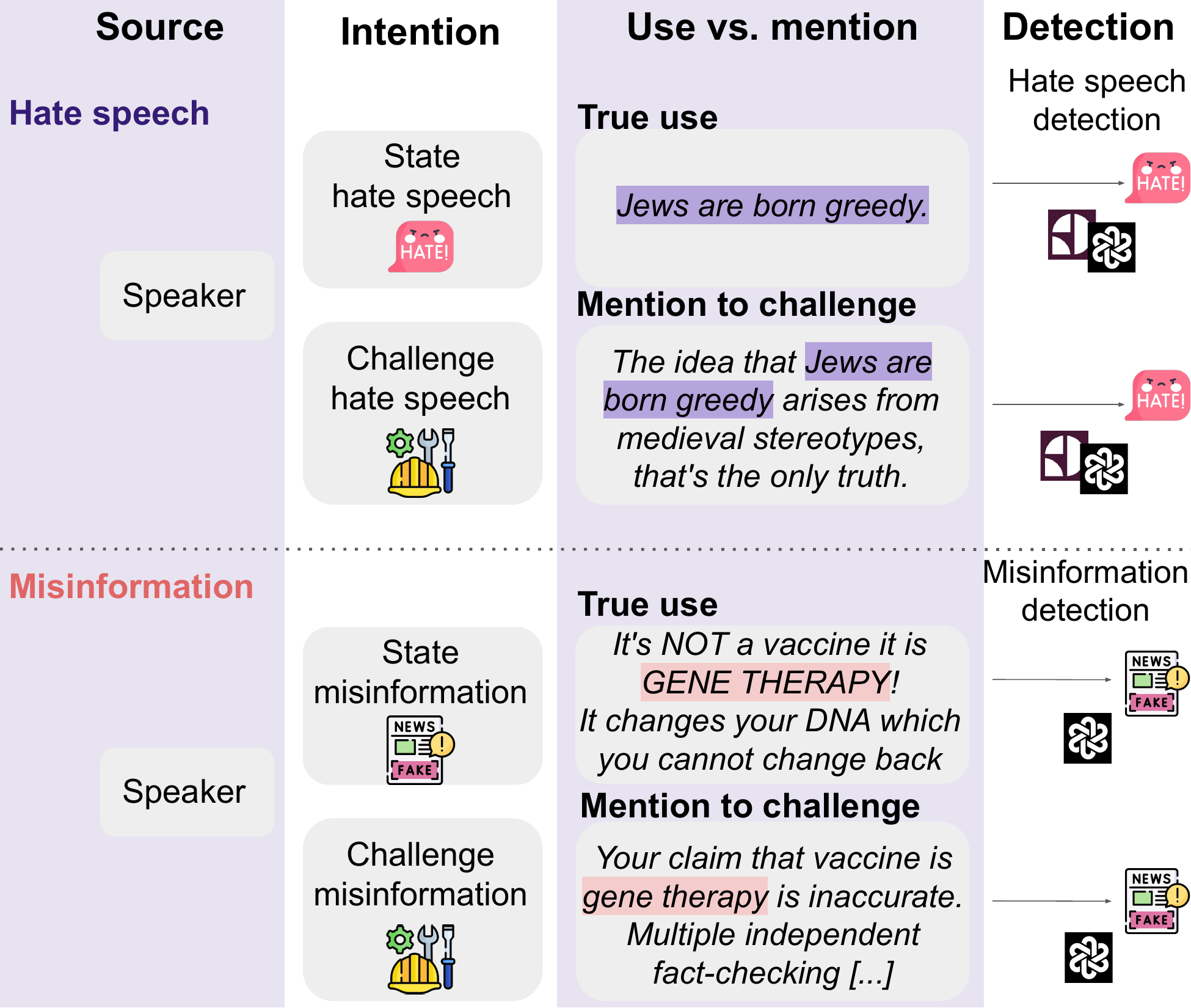}
    \caption{\textbf{Illustration of harmful censorship of counterspeech.} Depending on the speaker's intention, the same phrases are either \emph{used} to state hate speech and misinformation, or \emph{mentioned}, to provide a counternarrative. Algorithms that indiscriminately classify counterspeech as abusive censor constructive discourse.}
\label{fig:diagram}
\end{figure}

\begin{table*}[ht]
\small
\centering
\begin{tabular}{p{0.48\textwidth} | p{.48\textwidth}  }
\toprule
\textbf{Hate speech statements, use \boldmath{$\{u_t\}$}} & \textbf{Counterspeech, mention \boldmath{$\{m_t\}$}} \\
\midrule

Our children are \hlterciarytab{\textbf{being forced fed}} with this 'Islam' nonsense in schools [...]  & What do you mean by \hlterciarytab{\textbf{being forced fed}}? I haven't seen anything like this.\\
\midrule
\hlterciarytab{\textbf{Jews are born greedy}} & The charge that \hlterciarytab{\textbf{Jews are born greedy}} arises from medieval stereotypes and has nothing to do with the actual financial practices of Jews. \\
\midrule
%Jews can operate {\textbf{\hlterciarytab{with impunity, free from} \hlterciarytab{\mbox{criticism}}}}. &  I don't see how you can say that Jews are operating "{\hlterciarytab{\textbf{with impunity, free from}} \textbf{ \hlterciarytab{\mbox{criticism}}}}."  \\
I'm tired of woman taking part to political debate. \hlterciarytab{\textbf{Feminism is a divisive force in society}}. &  Of course, if you’ve never read anything about feminism except what its detractors purport that it’s about, then you come to believe \hlterciarytab{\textbf{feminism is a divisive force for society}}. But if you open yourself to different point of view you will change your mind, trust me! \\

\toprule
\textbf{Misinformation statements, use \boldmath{$\{u_t\}$}} & \textbf{Counterspeech, mention \boldmath{$\{m_t\}$}} \\
\midrule

My new book says, ``a vaccine, while not impossible, may be far off.'' Then people say, ``Wait Jim, there is a vaccine.'' Actually, no. What we have is \hlsecondarytab{\textbf{an experimental gene therapy}}. The mRNA shot is something like chemotherapy for people who don't have cancer. & The claim that the vaccine is \hlsecondarytab{\textbf{an experimental gene therapy}} is inaccurate. The vaccine only uses mRNA for immunization goal. It does not intertact with the gene at any levels. Please do not share wrong information. \\
\midrule
Chief Justice Mogoeng of South Africa calls COVID vaccine ``satanic ,'' a \hlsecondarytab{\textbf{plot to alter DNA}} to inscribe mark of Antichrist on everyone. These delusions may start in US, they spread through the world  & ``The \hlsecondarytab{\textbf{plot to alter DNA}}'' is wrong. As mentioned by CDC, the vaccine only uses mRNA, and do not interact with DNA. The Chief is misquoted by something, You should convey true information.  \\
\bottomrule
\end{tabular}
\caption{\textbf{Examples of statements using problematic language, and paired statements mentioning it.} Focal phrases are highlighted (used on the left, and mentioned on the right).}
\label{tab:examples}
\end{table*}

The ability to correctly make this distinction is particularly relevant to dealing with problematic language online. While detecting problematic text has become a standard task in NLP, there has been less consideration of text that might \textbf{mention} harmful content without directly \textbf{using} it. Mentioning problematic content is critical to language like
counterspeech that challenges or opposes hateful or misleading narratives (Table~\ref{tab:examples}) ~\cite{wright_vectors_2017,mun_beyond_2023,hangartner_empathy-based_2021,ecker_psychological_2022}. And mentioning is similarly crucial in {media and academic reporting} where researchers and journalists report harmful content~\cite{kirk_handling_2022}, in {educational settings} where problematic material is invoked for educational purposes, in disclosures in legal settings where harmful statements need to be quoted~\cite{henderson_pile_2022}, and in personal testimonies~\cite{wexler__2019}.

In this work, we focus on the first of these: \textbf{online counterspeech}, defined as speech produced by users of online platforms to counteract harmful speech of others. This effort seeks to stop the spread of harmful speech, mitigate its effects, discourage its recurrence, and provide support to both the targeted individuals and those joining in the counterspeech efforts~\cite{garland2022impact}.  

Counterspeech statements often involve referring to or quoting problematic content~\cite{vidgen_learning_2021}. Errors in distinguishing use from mention might therefore lead to failures in downstream classification, making counterspeech statements more likely to be misclassified as harmful by modern NLP systems, as shown in Figure~\ref{fig:diagram}.
Yet counterspeech helps curb online abuse \cite{bonaldi_human-machine_2022} and make online spaces safer~\cite{siegel_no2sectarianism_2020}. Thus, erroneously classifying counterspeech as problematic leads to content removal with significant implications: misclassification erases opportunities to rectify false narratives, and in doing so, risks further censoring those already most affected by harmful language~\cite{sahoo_detecting_2022,rahman_n_2012,park_reducing_2018}.  

But addressing the use-mention distinction and assessing its impact on downstream tasks is challenging. First, reasoning about the distinction itself has been hard, due to the lack of datasets, resources, and quantitative measurement methods. The few prior studies have been small and limited to linguistic features like particular mention verbs \cite{wilson_search_2011}.
Second, online counterspeech generally occurs in informal contexts, where markings that formally indicate mention, such as quotation marks or italics, are often missing~\cite{wilson_computational_2011}.
Finally, since mentioned language is less frequent than use~\cite{wilson_search_2011}, it is easy for researchers to overlook the downstream performance of NLP systems on mentioned language.

Motivated by these technical challenges, failure cases in counterspeech (Figure~\ref{fig:diagram} and Table \ref{tab:examples}), and literature on lexical and topic biases in harmful content detection~\cite{dixon_measuring_2018,ethayarajh_understanding_2022}, we make the following hypotheses:
\paragraph{H1} NLP models fail to distinguish use from mention in counterspeech.
\paragraph{H2} Failure in distinguishing use from mention impacts downstream tasks like hate speech and misinformation detection.
\paragraph{H3}  What is considered `permissible' by hate speech or misinformation classifiers is influenced by the presence of identity terms and other targeted entities, as well as the strength of the stance expressed toward mentioned language.
\paragraph{H4} Downstream performance can be improved by teaching the use-mention distinction to {explicitly} encode the treatment of mentioned language. 

Testing these hypotheses, our contributions include:
%  suggest that NLP models may have trouble distinguishing the use from mention, leading to tangible harms like harmful censorship of counterspeech. In this work,

%\paragraph{Contributions} Focused on the domain of counterspeech against misinformation and hate speech (Table \ref{tab:examples}), we contribute:

\paragraph{(1) Use-mention tasks} We formalize two tasks with challenging use-mention examples: (1) use versus mention classification and (2) downstream hate speech and misinformation detection (Sec.\ref{sec:methods}). 

\paragraph{(2) Failure analyses} We identify use-mention distinction failures, show that errors propagate to downstream tasks, and trace failures to specific target entities (misinformation-related and identity terms) and to the strength of the expressed stance (Sec.\ref{sec:results}--\ref{sec:res3}). 

\paragraph{(3) Mitigations} We investigate prompting mitigations and implications for downstream tasks. We show that our interventions lead to a significant reduction in error (Sec.~\ref{sec:res4}). 

To support the further evaluation and development of models, our code is available at \href{https://github.com/kristinagligoric/use-mention}{https://github.com/kristinagligoric/use-mention}. Datasets are publicly available and can be used for research purposes.

\section{Background}\label{sec:background}

\paragraph{Counterspeech} 
Building upon social science literature showing that counternarratives are effective against hate speech \citep{andrews2002introduction,benesch2014countering,schieb2016governing,garland2022impact}, previous work in NLP and HCI has studied counterspeech from various perspectives. Scholars have curated datasets of counterspeech from different sources, including experts, NGO workers, and social media comments \citep{mathew2019thou, chung2019conan, qian_benchmark_2019,garland2020countering,fanton2021}. Others have investigated different types of counterspeech strategies and performed user studies to evaluate the effectiveness of machine-generated counterspeech \citep{mun_beyond_2023,fraser-etal-2023-makes}. Our work also builds upon existing models for counterspeech detection \citep{garland2020countering}.

\paragraph{Content moderation policies on counterspeech mentions}

Several online platforms acknowledge the importance of counterspeech in their content policies. At the time of writing, TikTok's Community Principles states ``\textit{We do not allow language or behavior that harasses, humiliates, threatens, or doxxes anyone. This also includes responding to such acts with retaliatory harassment (but excludes non-harassing counter speech)}''~\cite{tiktok_safety_2023}. Similarly, Facebook publisher and creator guidelines state ``\textit{We know that many publishers use Facebook to challenge ideas, institutions and practices. Such discussion can promote debate and greater understanding}''. The guidelines also provide advice on how to write counterspeech to avoid mislabeling as hate speech~\cite{meta_hate_2023}. Development of models that enable enforcement of such policies is thus pressing.

While previous work has not explored the impact that the use-mention distinction has on online text classification tasks, mentions of toxic phrases were described as one class of false positive errors in toxic comment classification~\cite{van_aken_challenges_2018}, e.g., ``\textit{I deleted the <identity group> are dumb comment}'', establishing that errors due to use-mention distinction are prevalent.

\paragraph{The use-mention distinction}

The use-mention distinction has been studied in philosophy~\cite{sperber_irony_1981}, computational linguistics ~\cite{wilson_distinguishing_2010,wilson_search_2011,behzad_elqa_2023}, and HCI \cite{anderson_use-mention_2002}. In general, mention is defined as follows:
\begin{definition}
``for a token or a set of tokens $t$ in a sentence $S$, if $t$ refers to a property of the token $t$, then $t$ is an instance of mention.''  \cite{wilson_distinguishing_2010}
\label{definition1}\end{definition}

While many facts about a token can be a mention property,
in our domain of counterspeech we focus on two properties of mentioned language~\cite{sandhan-etal-2023-evaluating,wilson_search_2011}: \textbf{attributed language} (e.g., mentions to refer to the quotes of original source or stance towards the source) and \textbf{words or phrases as themselves} (e.g., mentions of words to refer to stance towards their use).\footnote{See the Limitations section on other types of mentioned language which do not lead as clearly to practical harms.}

\paragraph{The use-mention distinction and related tasks}

Like the use-mention distinction, other NLP tasks bear on speaker intent, such as those related to factuality~\cite{sauri2012you,sauri2009factbank,murzaku2022re} and committed belief~\cite{prabhakaran2010automatic,prabhakaran2015new,de2019commitmentbank}. In the context of mentioned language, such tasks aim to directly take into account the speaker's intention when mentioning specific phrases, and in particular, whether the speech act commits the speaker to the truth or factuality of the expressed proposition. However, in both used and mentioned language, statements do not necessarily commit the speaker to the truth of the source statement or to its factuality. Moreover, harmful language is often implicit, formalized as questions and nuanced statements that need not constitute a committed belief or be factual either. The distinction between using and mentioning is thus related to but distinct from factuality and committed belief. 

Lastly, mentioned language is also an instance of metalanguage \cite{behzad_elqa_2023,perlis_conversational_1998,wilson_creation_2012,wilson_toward_2013}, and our task builds upon work on
similar tasks like distinguishing whether personal names are being used to mention or address  \cite{prabhakaran_distinguishing_2023}.

\section{Methods}\label{sec:methods}

\subsection{Tasks} We focus on language indicative of hateful speech or misinformation, two frequent and connected types of harmful online content~\cite{mosleh2024misinformation}, which can be used or mentioned to express disapproving attitude towards it (as illustrated in Fig \ref{fig:diagram}). We hypothesize that use-mention distinction failures cause harmful misclassifications of counterspeech on downstream tasks. To test this hypothesis, we operationalize two tasks: the use-mention distinction and downstream classification. 

\paragraph{Task 1: Use-mention classification} For a given text, the task is to classify whether hateful/misinformative language is used or whether it is mentioned. True use $U$ are statements which use hate or misinformation, and $M$ are counterspeech mentions. For each text $u\in U$, $m\in M$, we classify the text as either use (positive class) or mention (negative class). Metrics we report are \textbf{false positive rate} and \textbf{false negative rate} in detecting use, and \textbf{average error rate}, capturing the average of the two rates. False positives are mentions misclassified as uses, while false negatives are defined as uses misclassified as mentions. On Task 1, false positives (mistaking mention for use) are the errors of primary interest due to their hypothesized impact on downstream tasks. 

\paragraph{Task 2: Downstream classification where use-mention distinction matters} We address two important downstream sub-tasks: hate speech detection and misinformation detection, with challenging use-mention examples. Similarly, for each text $u\in U$, $m\in M$, we classify each statement as either the positive (``misinformation'' and ``hate speech'') or negative class (``not misinformation'' and ``not hate'') on the downstream task. Since the standard metric is false positive rate capturing how often non-harmful content is misclassified as harmful~\cite{dixon_measuring_2018,markov2021improving}, we report \textbf{false positive rate} (mentions misclassified as ``misinformation'' or ``hate''). We also report \textbf{false negative rate} (uses misclassified as ``not misinformation'' or ``not hate''), and \textbf{average error rate}, capturing the average of the two rates. An ideal model would classify all $u_i$ as ``misinformation'' or ``hate speech'' (0\% false negative rate), while all counterspeech statements $m_{i}$ would be classified as ``not misinformation'' or ``not hate'' (0\% false positive rate). On Task 2, false positive rate on $M$ is the pragmatic concern central to our investigation as it captures the censorship rate.

\begin{table*}[t]
\centering
\small
%\begin{tabular}{ll|r|r|r}
\begin{tabular}{p{0.12\textwidth} p{0.23\textwidth} | R{0.17\textwidth}  |R{0.17\textwidth} | R{0.18\textwidth}}
\toprule
\textbf{Sub-task} &\textbf{ Model}  & \textbf{False positive rate {$\downarrow$}} & \textbf{False negative rate {$\downarrow$}} &  \textbf{Average error rate {$\downarrow$}}\\
\midrule
\textbf{Hate speech} &  gpt-3.5-instruct-turbo &           17.98 $\pm$          7.55 &   14.77 $\pm$         6.73 & 16.38 $\pm$  5.34 \\
 &          gpt-3.5-turbo (ChatGPT 3.5)  &       6.82 $\pm$           4.37 &   20.00 $\pm$         6.38 &     13.48 $\pm$            3.91 \\
 &            gpt-4 &      20.00 $\pm$          8.57 &    4.44 $\pm$       4.34  &     \textbf{12.22} $\pm$         4.74\\

\midrule
\textbf{Misinformation} &  gpt-3.5-instruct-turbo  &      34.38 $\pm$            3.76 &   40.08 $\pm$          3.33 &     37.22 $\pm$            2.21\\
 &          gpt-3.5-turbo (ChatGPT 3.5)  &       8.05 $\pm$            1.83 &   49.76 $\pm$         3.17 &     28.93 $\pm$            1.86\\
 &           gpt-4  &      23.44 $\pm$             2.74 &    3.89 $\pm$          1.38 &     \textbf{13.64} $\pm$            1.42\\
\bottomrule
\end{tabular}
\caption{\textbf{Use-mention classification.} False positive, false negative, and average error rates in detecting use (in percentages). Models are sorted by the average error rate. The best performing model has a high false positive rate, mistaking mention for use, which can lead to censorship of useful counterspeech.}
\label{tab1}
\end{table*}

\begin{table*}[t]
\small
\centering
\begin{tabular}{p{0.12\textwidth} p{0.23\textwidth} | R{0.17\textwidth}  |R{0.17\textwidth} | R{0.18\textwidth}}\\
\toprule
\textbf{Sub-task} &  \textbf{Model } &  \textbf{False positive rate {$\downarrow$}} &  \textbf{False negative rate {$\downarrow$}} & \textbf{ Average error rate {$\downarrow$}}  \\
\midrule
\textbf{Hate speech} &                        toxigen hatebert   &      24.44 $\pm$            10.0 &      77.78 $\pm$            7.22 &      51.11 $\pm$     8.08 \\
 &                                 perspective (insult) &       4.44 $\pm$             3.89 &     61.11 $\pm$          9.85  &      32.78 $\pm$    6.27 \\
  &                               perspective (toxicity)  &      20.00 $\pm$             7.51 &      36.67 $\pm$             9.18 &       28.33 $\pm$     8.49 \\
 &                        perspective (identity attack)  &      21.11 $\pm$            7.51 &       33.33 $\pm$             9.74&      27.22 $\pm$    9.18 \\

 &  roberta hate speech  &      17.78 $\pm$             7.51  &      26.67 $\pm$             8.65 &       22.22 $\pm$     8.22 \\
  &                               gpt-3.5-instruct-turbo &      25.56 $\pm$             8.62 &      13.33 $\pm$             7.51 &       19.44 $\pm$    8.20 \\
 &                                       gpt-3.5-turbo (ChatGPT 3.5) &      11.11 $\pm$           6.14 &       22.22 $\pm$           8.07  &    16.67 $\pm$     7.24 \\

 &                                         gpt-4 &             8.89 $\pm$          5.00 &  20.00 $\pm$             7.51 &       \textbf{14.44} $\pm$    7.09\\

\midrule

 \textbf{Misinformation} 
&   roberta fake news   &      97.93 $\pm$             1.00 &        5.10 $\pm$             1.52  &  51.52 $\pm$    1.20 \\
 &                           gpt-3.5-instruct-turbo &            26.12 $\pm$             2.40 & 19.44 $\pm$             2.62 &        22.78 $\pm$    2.80 \\
 &                                   gpt-3.5-turbo (ChatGPT 3.5) &          22.11 $\pm$             3.02  & 13.85 $\pm$             2.74 &        17.98 $\pm$     3.06 \\
 &                                     gpt-4 &           10.21 $\pm$             2.00 & 8.02 $\pm$             1.85              &       \textbf{9.11} $\pm$     1.93 \\

\bottomrule
\end{tabular}
\caption{\textbf{Downstream tasks (hate speech and misinformation detection).} False positive, false negative, and average error rates in detecting hate speech and misinformation (in percentages). Models are sorted by the average error rate. Across models, substantial errors persist in false positive rate (classifying counterspeech mentions as harmful).}
\label{tab:downstream}
\end{table*}

\subsection{Datasets}

\paragraph{Countering hate} For this task we rely on two datasets: Knowledge-grounded hate countering \cite{chung_towards_2021} and Multi-Target Counternarratives \cite{fanton2021}. The datasets contain pairs of (hateful statement, counterspeech), illustrated in Table~\ref{tab:examples}. We select counterspeech statements written by human experts. 

\paragraph{Countering misinformation} For this task we leverage misinformation counternarratives \cite{he_reinforcement_2023}. The dataset contains pairs of (misinformative statement, counterspeech), illustrated in Table~\ref{tab:examples}. Misinformative statements were posted on social media, while counter-responses are a mix of naturally occurring social media posts and counterspeech statements written by recruited human participants. 

\paragraph{Focal tokens} To confirm that mentioned language is indeed relevant to the practical case of counterspeech, we verified that counterspeech mentions contain language from the original true use sample it addresses, which we refer to as focal tokens. Across $N=1826$ pairs $\{(u_t,m_t)\}$, we computed the length of the longest common substring using a dynamic programming algorithm \cite{suzgun_string2string_2023}. We found substantial overlap (as in examples in Table~\ref{tab:examples}, focal tokens), with average $M=3.44$ words in the longest common substring. Additionally, for both datasets, we manually verified that focal tokens are used in true use statements (and not mentioned), and that counterspeech is not using, but mentioning (see Appendix, Sec.~\ref{dataset} for details). 

\subsection{Models} For the use-mention task, we tested the two best performing GPT models at the time of writing (gpt-3.5-turbo, gpt-4), as well as gpt-3.5 instruct, the non-RLHF legacy variant, using zero-shot prompting. For downstream tasks, we tested these three models as well as four widely-used models for hate speech and misinformation detection: Perspective~\cite{perspective_using_2023}, Toxigen~\cite{hartvigsen_toxigen_2022}, RoBERTa~\cite{liu_roberta_2019}, 
and RoBERTa fake news~\cite{ahmed_detection_2017}.

For prompting, we use default parameters (temperature=1) and max output token length 1 (outputting either A or B). The classification prompt includes the instruction and a definition of the classes (for prompt variants and complete prompt text see Appendix, Section~\ref{sec:prompts}).

\section{Results}\label{sec:results}

\subsection{Use-mention classification (H1)}\label{sec:res1}

How well do the models distinguish whether problematic language is used or mentioned? Across the models (Table~\ref{tab1}), average error rates are high (between 12.22\% and 16.38\% for hate speech and between 13.64\% and 37.22\% for misinformation), suggesting that state-of-the-art large language models struggle to distinguish use from mention in domains where the distinction matters. The best-performing model, gpt-4, still has a very high false positive rate, mistaking mention for use; mentions are misclassified as use in \textbf{{20.00\%}} of hate speech and \textbf{{23.44\%}} of misinformation counternarratives. In these settings, mistaking mention for use leads to the consequential harm of censoring useful counterspeech. 
% so these gpt-4 errors are indeed a harmful type of error. 

\subsection{Downstream content classification (H2)}\label{sec:res2}

\begin{table*}[t]
\small
\centering
\begin{tabular}{ p{0.23\textwidth} | R{0.19\textwidth} | R{0.19\textwidth}| R{0.08\textwidth}| R{0.1\textwidth} }
\toprule
                       \textbf{Model} &  \textbf{Use-mention $\mathbf{\neg}$correct}    & \textbf{Use-mention correct}    & $\mathbf{{\chi}^2 }$              & $\mathbf{p}$                                         \\
\midrule
gpt-3.5-instruct-turbo      &     \textbf{32.96\% }                          &                 19.52\%                 &      20.58      &    $5.72\times10^{-6}$       \\
 gpt-3.5-turbo (ChatGPT 3.5) &          \textbf{28.31\% }                     &  14.44\%                                &      25.08       & $5.51\times10^{-7}$           \\
 gpt-4                       &          \textbf{15.78\%}                     &    4.54\%                               &      30.60       &  $3.17\times10^{-8}$ \\

\bottomrule
\end{tabular}
\caption{\textbf{Error propagation.} False positive rate in downstream classification of counterspeech mentions, stratified by use-mention classification correctness. Error rates on counterspeech are significantly higher when use-mention distinction is incorrect. Statistics for hate speech and misinformation separately are listed in the Appendix, Table~\ref{tab:propagation_by_task}.}
\label{tab:propagation}
\end{table*}

The prior section showed that state-of-the-art systems often fail at the use-mention distinction. Here we test the impact on hate speech and misinformation detection, two downstream classification tasks where the use-mention distinction matters (Table~\ref{tab:downstream}).

\paragraph{Performance on downstream tasks} First, on the hate speech detection task, we find that  misclassification of counterspeech as hateful is relatively frequent: the popular Toxigen and Perspective API have a false positive rate on counterspeech of \textbf{over 20\%}, and recent models still have many false positives, although somewhat reduced, e.g., gpt-3.5-turbo has FPR 11.11\%. Regarding the average error rate, gpt-4 is the best-performing model (false positive rate on counterspeech~\textbf{8.89\%}).

Second, on the misinformation detection task, misclassification of counterspeech as misinformative is relatively frequent: gpt-3.5 models have \textbf{over 20\%} false positive rate on counterspeech, and substantial errors persist even with the best system, gpt-4 (\textbf{10.21\%} false positive rate classifying counterspeech mentions as misinformation). We also note that two investigated models, Togixen and RoBERTa fake news, have average error rates above 50\%.

\paragraph{Error propagation to downstream tasks}

As a further test of the hypothesis that errors in distinguishing use from mention  cause these failures in downstream tasks, we assess whether examples that cause errors  downstream (Task 2) are also misclassified in   use-mention classification (Task 1). We do this by testing how the downstream error rate on Task 2 (hate speech detection and misinformation detection) associates with the error rate on Task 1 (use vs. mention). For gpt-4, for example, we contrast the error rate on Task 2 of 15.78\% (aggregated over hate speech and misinformation detection) for statements in which the Task 1 classifier failed at distinguishing use from mention, versus the error on Task 2 of 4.54\% (aggregated) for statements in which the Task 1 classifier correctly distinguished use from mention, comparing these rates using the chi-squared test.

We find that, among samples where mentioning counterspeech is misclassified as use, downstream misclassification is   higher across the three tested models (all $p<10^{-5}$; Table~\ref{tab:propagation}). Results disaggregated by hate speech and misinformation detection are listed in the Appendix (Table~\ref{tab:propagation_by_task}).

In summary, we find that errors in use-mention distinction do propagate to downstream tasks. %Among mentioning counterspeech \emph{incorrectly} classified as use, the error rate on downstream tasks is higher than among examples \emph{correctly} classified as mentioned language. 
%The fact that errors in use-mention distinction propagate to downstream tasks underscores the importance of the distinction. 

\subsection{Why is the distinction hard (H3)?}\label{sec:res3}

%Although counterspeech that opposes harmful narratives is bynot harmful by definition, in downstream tasks, NLP tools make judgments regarding when mentioned language in counterspeech is permissible vs. not. Next, we aim to understand the factors that guide this process.
Counterspeech that opposes harmful narratives is by definition not harmful. What linguistic aspects of this counterspeech cause downstream NLP tools to incorrectly label counterspeech as harmful? Or, viewed from the other perspective, what terms do NLP tools make permissible as mentions?

\paragraph{Hate speech: Target identity terms}

We know that
toxicity detection algorithms rely heavily on the presence of identity mentions to make their predictions \cite{zhou_challenges_2021}.
For the hate speech task, we therefore hypothesize that identity terms will impact when mentioned language is permissible. To test this hypothesis, we stratify error rate in classifying counterspeech as hate speech by target identity (as labeled in the metadata of the dataset~\cite{chung_towards_2021}).

We find that errors vary widely depending on the targeted identity (e.g., gpt-3.5-turbo: Jewish 14.15\%, people of color 9.09\%, Muslims 6.80\%, LGBT+ 6.77\%, while for the other groups it is less than 5\%; Table~\ref{tab:bytarget}). Even the most recent large language models' treatment of counterspeech similarly varies depending on the mentioned identity. Our results suggest that systems treat the mere mention of certain identity terms as impermissible (see Sec.\ref{sec:discussion}).

\begin{table}[t]
\small
\centering
\begin{tabular}{p{0.3\columnwidth} | R{.25\columnwidth} | R{.25\columnwidth}  }
\toprule
\textbf{Target identity} & \textbf{gpt-3.5-turbo } & \textbf{gpt-4} \\
\midrule
Jewish       & 14.15\% & 12.15\%\\
People of color     &     9.09\% & 9.09\% \\
Muslims   &   6.80\% & 4.83\%\\
LGBT+     &   6.77\% & 2.22\%\\
Disabled  &   4.00\% & 0.00\%\\
Women   &     2.41\% & 1.18\%\\
Other   &    0.89\% & 1.77\% \\
Migrants  &   0.39\% & 0.76\%\\
\bottomrule
\end{tabular}
\caption{\textbf{Identity.} False positive rate in hate speech classification of counterspeech, stratified by target identity.}
\label{tab:bytarget}
\end{table}

\begin{table*}[ht]
\centering
\small
    \begin{tabular}{p{0.045\linewidth}| p{0.45\linewidth}|p{0.41\linewidth}}
   \textbf{Model }&  \textbf{Top terms for} \hlsecondarytab{\dno} & \textbf{Top terms for }\hlprimarytab{\dyes} \\
    \toprule  

\textbf{gpt- 4} & {please (-3.67), vaccine (-3.61), therapy (-3.53), gene (-3.32), misinformation (-3.23), stop (-2.82), mrna (-2.52), dna (-2.43), uses (-2.41), wrong (-2.27), change (-2.10), correct (-2.03), safe (-2.01), cdc (-1.99)}, well (-1.93)
    & {fake (5.03), news (4.89), lying (4.27), lies~(4.16), one (3.94), misleading (3.84), lie (3.76), put (3.43), thing (3.35), always (3.13), false (3.10), new (2.98), everything (2.83), anyway (2.83), country (2.83)}
\\
\midrule
\textbf{gpt- 3.5} & {mrna (-4.24), vaccine (-3.40), dna (-2.82), uses (-2.82), change (-2.77), please (-2.17), genes (-2.15), genetic (-2.12), actually (-2.09), gene (-2.06), therapy (-2.06), understand (-2.02)}, correct (-1.86), immunization (-1.86), nothing (-1.79)
    & {fake (5.39), misleading (5.08), lying (4.31), lies~(4.20), media (4.09), news (4.03), another~(3.60), report (3.39), journalism (3.37), blood~(2.94), false~(2.92), crazy (2.65), justice (2.64), remove~(2.48), reporting (2.45)}
\\
    \end{tabular}
    
    \caption{\textbf{Top 15 terms for \hlsecondarytab{\dno} and \hlprimarytab{\dyes}.} Z-scores (in parentheses) are computed using the Fightin' Words method. Words with $|\text{z-score}|>1.96$ are statistically significant in frequency difference. Misclassification of counterspeech mentions as harmful is influenced by specific COVID-19-related terms (``mRNA'', ``vaccine''), and strength in stance towards mentioned language (``fake'', ``misleading'').}

    \label{tab:topwords}
\end{table*}

\paragraph{Misinformation: COVID-19 terms and strength in stance towards the embedded language}

To understand why counterspeech mentioning misinformation is detected as misinformation, we use the Fightin’ Words method~\cite{monroe_fightinwords_2008} to measure statistically significant differences in tokens between two sets---counterspeech mentions classified as misinformation vs. not classified as misinformation---after controlling for variance in words’ frequencies.

We compute the top differentiating words for $D_{\checkmark}$ versus $D_{\times}$, where
\begin{equation*}
    \begin{split}D_{\checkmark} = \{m_i \in M | \text{misinformation}_c(m_i)=\text{True}\}\\
    \end{split}
\end{equation*}
\begin{equation*}
    \begin{split}
    D_{\times} = \{m_i \in M | \text{misinformation}_c(m_i)=\text{False}\}\\
    \end{split}
\end{equation*} 
respectively, where $M$ are all counterspeech statements and $\text{misinformation}_c$ is misinformation classification by classifier $c$.

We find that terms relating to controversial topics are often misclassified; top terms for $D_{\times}$ include  ``gene therapy'', ``mRNA'', ``vaccine'', ``DNA'', and ``CDC'', suggesting that counterspeech mentioning   COVID vaccination is often misclassified. Certain terms associated with COVID-19 misinformation are not permissible even when mentioned.

We hypothesize that due to interactions with safety features that prevent large language models from generating health disinformation~\cite{menz2024current}, the mere presence of specific terms related to COVID vaccination is linked with misclassification of counterspeech as misinformative. This finding that terms related to COVID are treated as `impermissible' when mentioned parallels our finding that mentions of certain demographic identities are impermissible to hate speech detectors.

We also find that top terms for $D_{\checkmark}$ are related to expressing a strong stance against misinformation in the surrounding context and the strength of meta language (as indicated by terms such as ``fake news'', ``lying'', ``lies'', ``misleading''). Downstream classification has fewer errors when the disagreement in mentioning statements is not subtle.

\paragraph{Distancing by using quotation marks}

Counterspeech that uses verbatim quotes typically involves more severe language, as quotes enable distancing~\cite{wilson_computational_2011}. We hypothesize that such texts are more censored.
Indeed, we find for both hate speech and misinformation that the counterspeech containing quotation marks is more frequently misclassified as harmful (Table \ref{tab:quotations}).

\paragraph{Summary of Error Analysis} In summary, misclassification of counterspeech mentions as harmful is influenced by over-reliance on (a) \textbf{surface terms} such as identity words and specific COVID-19-related terms, and (b) notions of \textbf{strength in stance towards mentioned language}.

\begin{table*}[t]
\small
\centering
\begin{tabular}{p{0.14\textwidth} | p{0.23\textwidth} | R{0.1\textwidth} | R{0.1\textwidth}| p{0.08\textwidth}| p{0.1\textwidth} }
\toprule
\textbf{Sub-task} & \textbf{Model} &  \textbf{Mention quotations }    & \textbf{$\mathbf{\neg}$Mention quotations}    & $\mathbf{{\chi}^2 }$              & $\mathbf{p}$                                         \\
\midrule
\textbf{Hate speech} & gpt-3.5-instruct-turbo      &        \textbf{57.14\%}                        &  22.89\%                               &       {3.98 }    &       0.046     \\

& gpt-3.5-turbo (ChatGPT 3.5) &             28.57\%                   &    9.88\%                              &     2.24        &       0.13                              \\
& gpt-4     &                28.57\%                 &         7.23\%                          &       3.63        &       0.056                               \\
\midrule

\textbf{Misinformation} & gpt-3.5-instruct-turbo      &        30.00\%                      & 26.12\%                                &      0.15     &      0.70    \\
& gpt-3.5-turbo (ChatGPT 3.5) &              \textbf{45.00\% }               &      21.79\%                            &      {6.05}      &     0.014    \\
& gpt-4                       &         \textbf{25.00\%}                     &     9.84\%                        &     {4.89 }       &  0.027\\
\bottomrule
\end{tabular}
\caption{\textbf{The impact of quotation marks.} False positive rate in downstream classification of counterspeech mentions, stratified
by the presence or absence of quotation marks. }
\label{tab:quotations}
\end{table*}

\subsection{How can we teach the distinction (H4)?}\label{sec:res4}

\begin{table*}[t]
\centering
\small
\begin{tabular}{p{0.10\textwidth} | p{0.09\textwidth} | R{0.15\textwidth} |R{0.17\textwidth} | R{0.15\textwidth} | R{0.17\textwidth}}
\toprule
    \textbf{Sub-task} &                                                \textbf{Mitigation} & \textbf{False positive rate (counterspeech)~{$\downarrow$} } &  \textbf{False positive rate $\Delta$ (counterspeech)~{$\downarrow$}  } & \textbf{True positive rate (true use) {$\uparrow$} } & \textbf{True positive~rate~$\Delta$ (true use) {$\uparrow$} }\\
\midrule
{\textbf{Hate speech }}  & \hlterciarytab{No mit.} & \hlterciarytab{8.89\%} & --- & \hlterciarytab{80.00\%} &  ---\\
       &                              Few shot & 5.02\% & -43.48\% & 79.81\% & -1.49\%  \\
        &                                      Mitigation & 5.41\% &  -39.13\%  & 79.81\%  & -1.49\%\\
       &                      CoT$+$mit. & \textbf{1.55\%} & {\textbf{-82.61\%}}  & 77.61\% & -2.99\% \\
      \midrule
    \textbf{Mis-}  &  \hlterciarytab{No mit.} & \hlterciarytab{10.21\%} & --- & \hlterciarytab{91.98\%} & ---\\
   \textbf{information}   &                             Few shot  & 5.28\%& -48.32\%  & 86.09\% & -6.40\% \\
     &                                   Mitigation & 7.33\%& -28.19\%  & 89.57\% &  -2.62\% \\
    &                         CoT$+$mit. & \textbf{4.18\%}&  {\textbf{-59.06\%}}  & 89.57\% & -2.62\%\\
\bottomrule
\end{tabular}

\caption{\textbf{Mitigations.} For each prompting mitigation, false positive rate, true positive rate, and relative change in the rates calculated as $\Delta=(\text{rate} -  \text{rate no mitigation})/ \text{rate}$. No mitigation rates are listed for reference. In Figure~\ref{fig:2d_m}, we illustrate the tradeoff between true positive rate (on true use) and false positive rate (on counterspeech mentions) across the tested models. Mitigation reduces false positive rate on counterspeech, with a small decrease in true positive rate for true use statements. Statistics are reported for gpt-4. For gpt-3.5-turbo, see Appendix~\ref{app:mitigations}.}
\label{tab:mitigations4}
\end{table*}

\paragraph{Methods} Informed by the analyses revealing that errors propagate from the inability to distinguish use from mention into misclassification on the downstream tasks, we explore a set of prompting mitigations to reduce downstream mistakes. In particular, we explore ways to teach the use-mention distinction through controlled prompting. 

To that end, we designed and tested CoT prompting mitigation. In \texttt{CoT $+$ mitigation}, we (1) embed the definition of use-mention distinction and an instruction specifying that mention of hateful or misinformative language does not imply that the text is hateful or misinformative. We use prompt formats inspired by BigBench CoT \cite{suzgun_challenging_2023} and (2) follow the process prompting the LLM to ``\textit{think step-by-step}''~\cite{wei_chain--thought_2022} and use the answer extraction prompt ``\textit{so the answer is}'' to the generated rationale to extract the final answer. We also (3) include few-shot examples of mentioned and used language where the first step considers whether potentially problematic language is used or mentioned, before making the classification in the second step (for complete prompt text, see Appendix, Table~\ref{tab:prompts_mitigations}).

Furthermore, we perform an ablation study isolating the impact of use-mention examples alone (\texttt{Few shot}) and embedding the instruction to make the use-mention distinction alone (\texttt{Mitigation}).

For the best-performing model (gpt-4), we tested the prompting mitigation on counterspeech mentions and true uses. Intuitively, the mitigation should reduce the misclassification of counterspeech mentions (false positive rate on counterspeech) while not reducing correct positive classification of true uses (true positive rate on uses).

\paragraph{Results} We find that the CoT mitigation reduces false positive rate among counterspeech mentions by 82.61\% for hate speech and 59.06\% for misinformation (Table~\ref{tab:mitigations4}). Among true use statements, CoT mitigation reduces true positive rates only marginally (2.99\% for hate speech and 2.62\% for misinformation). Ablated mitigations reduce false positive rate among counterspeech mentions independently, but \texttt{CoT + mitigation} is the best performing condition. With gpt-3.5-turbo, similar patterns were observed (see Appendix~\ref{app:mitigations}).

In summary, encoding the use-mention distinction reduces downstream misclassification with otherwise minimal reductions in performance on true use of hate speech and misinformation.

\section{Discussion}\label{sec:discussion}

\paragraph{Implications for content moderation} Existing datasets of harmful content are typically collected by sampling keywords that co-occur with harmful content and performing annotation. For example, during hate speech annotation, a typical question might ask: ``Does the above text contain rude, hateful, aggressive, disrespectful, or unreasonable language?''~\cite{rae2021scaling}. Previous work has documented that existing datasets gathered through such a process contain mentioned language misclassified as harmful~\cite{van_aken_challenges_2018}. This implies that researchers may not typically consider the special case of mentioning statements (including counterspeech) when collecting annotations, or that annotators with varying backgrounds and positionality may not agree on how to treat mentioning statements~\cite{santy_nlpositionality_2023}.

By classifying hate speech and misinformation, NLP tools make \emph{implicit} judgments regarding when mentioned language is permissible. However, counterspeech should be permissible by definition, as it challenges and opposes harmful narratives~\cite{mun_beyond_2023,hangartner_empathy-based_2021}. 
We suggest that in downstream content classification, treatment of mentioned language should instead be \emph{explicitly} encoded. More broadly, efforts to use LLMs for the design of conversational socio-technical systems should take into account the use-mention distinction and strive to embed values regarding how content should be treated. To that end, our prompting mitigation serves as an example of how one societal value can be encoded by leveraging a linguistic construct to specify the treatment of counterspeech. 

\paragraph{Beyond surface features} A substantial body of literature has examined shortcomings of online content classification~\cite{garg_handling_2023,van_aken_challenges_2018}. Our results suggest that many such error cases are linked to the inability to sufficiently distinguish between use and mention.

For instance, previous work has suggested that the mere fact that a text contains \textbf{swear words}~\cite{ethayarajh_understanding_2022}, \textbf{identity terms}~\cite{dixon_measuring_2018} related to ethnicity~\cite{ghosh_detecting_2021}, religion~\cite{sheth_defining_2022}, gender, race, and disability~\cite{dias_oliva_fighting_2021,diaz_double_2021}, or \textbf{dialect markers} \cite{sap_risk_2019,halevy_mitigating_2021} increases toxic classifications~\cite{zhou_challenges_2021}. Such lexical biases where surface features spuriously influence classification are indicative of inadequate ability to make a use-mention distinction. 

Surprisingly, we find that even the latest models (i.e., gpt-4) are very sensitive to shallow lexical biases. Although non-literal language understanding beyond surface features is essential to human communication, large language models tend to struggle with interpretations of subtle utterances \cite{hu_fine-grained_2023,ocampo_-depth_2023,yu_hate_2022}. Our work thus extends prior work aiming to perform more nuanced online content classification~\cite{pamungkas_you_2020,goyal_is_2022,sap_social_2020}. Future work needs to attend to these subtle and implicit meanings to keep misinformation and hate speech classifiers from being mere surface topic classifiers.

\paragraph{Beyond counterspeech}\label{sec:future}

While we focus on online speech, the use-mention distinction might be important in other contexts like education or tutoring (where mention language occurs frequently to specify spelling or translation), law \cite{henderson_pile_2022}, and human-AI interaction~\cite{shaikh_grounding_2023}. In general, our work offers promising directions for embedding meta-linguistic reasoning to improve performance on challenging downstream tasks, such as eliciting metaphorical meanings, or dog whistle classifications, where existing methods are not reliable \cite{wachowiak_does_2023,mendelsohn_dogwhistles_2023} and teaching meta-linguistic skills might offer a possible way forward.

\section{Conclusion}\label{sec:conclusion}

This work highlights the theoretical and practical importance of the use-mention distinction in NLP and CSS. We also provide guidelines and directions for future research in modeling mentioned language and mitigating the impact on downstream tasks where failure to distinguish mention leads to harmful misclassifications.

\section*{Ethical implications}\label{sec:ethics}

The central implication of our work is that downstream tasks in applications that involve occurrences of mentioned language should be handled with caution, especially when misclassification of mention could be harmful. However, content categorized as counterspeech might not be universally beneficial. For example, humans can demonstrate bias in use of counterspeech to preferentially challenge content from those with whom they disagree politically~\cite{allen_birds_2022}. In that way, counterspeech can be leveraged as a tool to harass others with opposing views. Consequently, our mitigation to prevent censorship of counterspeech might allow harmful content to proliferate due to taking a form of counterspeech. We note that incorporating specific guidelines into platforms requires further testing. 

We also note that strategies involving mentioned language can equally be used to veil the speaker's true intention. Using specific terms might be socially acceptable for a person of a given identity, while unacceptable for someone else. For instance, acceptability of mentions of certain slurs is debated, especially if a non-derogatory version is readily available~\cite{green_beyond_nodate}. Addressing such complexities in the mentioned language largely remains an open question. 

Lastly, despite known limitations, natural language systems are widely used for online content moderation \cite{welbl_challenges_2021,gehman_realtoxicityprompts_2020}. It remains unclear how well they are expected to perform and whether it is appropriate to deploy models that might produce harmful errors \cite{fortuna_directions_2022}. Our work echoes the need for more cautious development and deployment of such systems that accounts for the conversational context, identities, and intentions. Our work also builds on existing literature that challenges the very concepts of binary classification~\cite{pachinger_toward_2023,davani_dealing_2022} into constructs such as hate speech. 

\section*{Limitations}\label{sec:limitations}

We limit the scope of our work to testing publicly available out-of the box classifiers. Similarly, we do not investigate all the possible mitigation strategies. For example, fine-tuning with more examples could help further decrease the error rates. We also note that we do not take into account how humans would rate studied texts. Previous work studying online discourse has found that people too struggle in disentangling mentioning a fact from stating an opinion, which makes the subsequent conversation more likely to derail into uncivil behavior \cite{chang_dont_2020}. Nonetheless, counterspeech examples examined in this work were written by social media users and expert humans who deemed them appropriate for responding to harmful narratives.

Finally, the present study is limited to specific types of mentioned language related to counterspeech. We do not solve all the problems of use-mention in the broader linguistics literature~\cite{sandhan-etal-2023-evaluating,wilson_search_2011}, including attributed language, words or phrases as themselves, proper names, and translations and transliterations. In this work, we are primarily interested in attributed language and words or phrases as themselves, as they closely relate to counterspeech which needs to mention phrases stated by others. Future work should consider other forms of mentioned language, and their impact on NLP systems more broadly. For instance, grammatical error correction might be impacted by failures in use-mention distinction~\cite{arand_grammarly_2022}. Our work is a first step toward analyzing the language of mention in different tasks, contexts, languages, cultures, and times.

\section*{Acknowledgments}
Kristina Gligorić is supported by Swiss National Science Foundation (Grant P500PT-211127). Myra Cheng is supported by an
NSF Graduate Research Fellowship (Grant DGE2146755) and Stanford Knight-Hennessy Scholars
graduate fellowship. This work is also funded by the Hoffman–Yee Research Grants
Program and the Stanford Institute for Human-Centered Artificial Intelligence.

\bibliography{anthology,refs}

\appendix

\section{Dataset statistics}\label{dataset}

Two annotators annotated a random sample of statements (160), uniformly distributed across hate speech and misinformation. Half of the instances were uses, and half counterspeech mentions. Each statement was annotated to indicate if the focal tokens are used or mentioned (following Def.~\ref{definition1}). Given that we leveraged datasets with curated counterspeech statements~\cite{fanton2021,chung_towards_2021,he_reinforcement_2023}, as expected, within the selected sample, true use original posts contained no mentions, while counterspeech contained no uses of hate speech and misinformation. Similarly, counterspeech mentions the same focal tokens from the original statement (as opposed to mentioning harmful language not in the original use statement).

\section{Mitigations}\label{app:mitigations}

\begin{figure*}[b]
\centering
\includegraphics[width=0.7\textwidth]{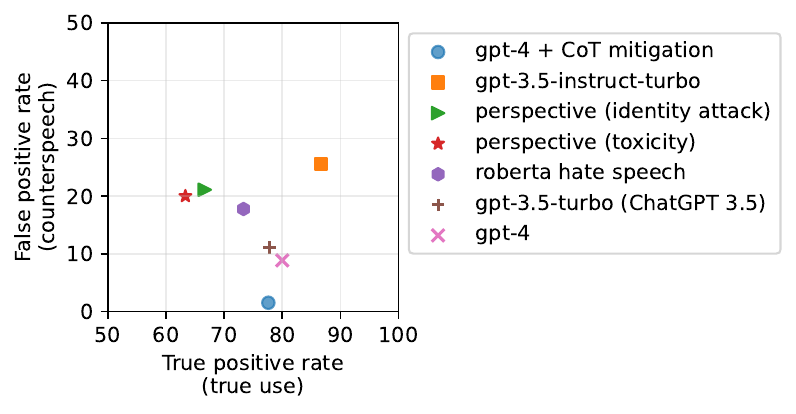}

\includegraphics[width=0.7\textwidth]{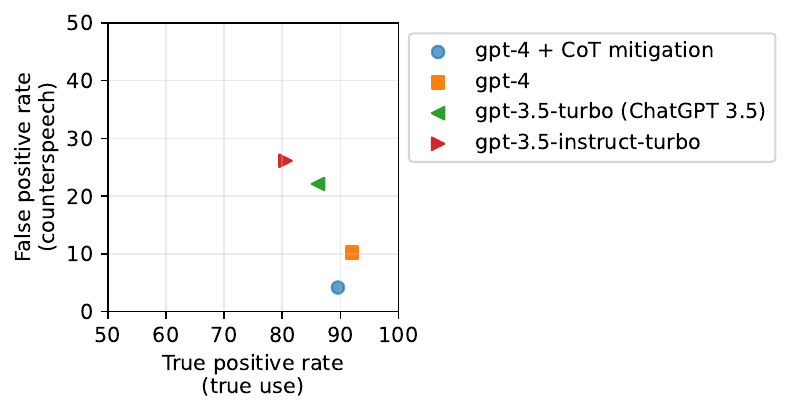}
\caption{\textbf{Mitigation illustration.} Metrics are visualized separately for hate speech detection (top) and misinformation detection (bottom). It is desirable to have a low counterspeech false positive rate (y-axis) and a high use true positive rate (x-axis). For both hate speech and misinformation, CoT mitigation reduces counterspeech false positive rate with a marginal decrease in true positive rate on use.}
\label{fig:2d_m}
\end{figure*}

In Figure~\ref{fig:2d_m}, we illustrate true positive rate on true use and false positive rate for counterspeech statements across the tested models. Mitigation reduces false positive rate on counterspeech, with a small decrease in true positive rate for true use statements. In Table~\ref{tab:mitigations_3.5}, for completeness, we list the results from mitigation study with gpt 3.5-turbo (ChatGPT 3.5). 

\section{Additional statistics}\label{sec:stats}

Table~\ref{tab:propagation_by_task} lists propagation statistics separately by task. Table~\ref{tab:recallbytarget} lists recall in hate speech classification stratified by target identity. We note that groups that have higher counterspeech false positive rates have a higher recall on true use, while groups with lower counterspeech false positive rates have lower recall too. Some discrepancies exist, as, for instance, false positive rate in mentioning counterspeech for Jewish is higher than for disabled identities, despite similar recall on true use (96\% and 98.11\%). These patterns are likely associated with biases in training data which implicitly encode the treatment of different groups, and the respective counterspeech.

\section{Prompt text}\label{sec:prompts}

For reproducibility, in Tables~\ref{tab:prompts} and \ref{tab:prompts_mitigations}, we list the complete prompts tested in our studies.

\begin{table*}[t]
\centering
\small
\begin{tabular}{l|l|r|r}
\toprule
    \textbf{Sub-task} &                                                   \textbf{Mitigation} &    \textbf{False positive }  & \textbf{True positive }\\
      &                                                          &    \textbf{ rate $\Delta$ {$\downarrow$}}  & \textbf{ rate $\Delta$} {$\uparrow$}\\
\midrule
       {\textbf{Hate speech}}  &        Few shot use-mention examples & -41.67\% & -2.94\%\\
         &               Use mention mitigation & -41.67\% & -5.88\%\\
       &  Few shot CoT use mention mitigation & {\textbf{-78.33\%}}  & -4.41\% \\
       \midrule
   \textbf{Misinformation}   &        Few shot use-mention examples & -58.02\% &  -7.74\%\\
      &               Use mention mitigation & -46.76\% & -6.50\% \\
     &  Few shot CoT use mention mitigation & {\textbf{-73.04\%}}  & -27.24\%\\
     
\bottomrule
\end{tabular}

\caption{\textbf{Mitigation statistics for gpt-3.5-turbo (ChatGPT 3.5)}.}
\label{tab:mitigations_3.5}
\end{table*}

\begin{table*}[t]
\small
\centering
\begin{tabular}{p{0.14\textwidth} | p{0.23\textwidth} | p{0.12\textwidth} | p{0.12\textwidth}| p{0.08\textwidth}| p{0.1\textwidth} }
\toprule
                       \textbf{Downstream task} & \textbf{Model} &  \textbf{Use-mention $\mathbf{\neg}$correct}    & \textbf{Use-mention correct}    & $\mathbf{{\chi}^2 }$              & $\mathbf{p}$                                         \\
\midrule
\textbf{Hate speech} & gpt-3.5-instruct-turbo      & \textbf{53.12\% }                              & 10.34\%                                 & {19.84}            & $8.42\times10^{-6}$           \\

& gpt-3.5-turbo (ChatGPT 3.5) & 19.35\%                               & 7.02\%                                 & 3.03             & $0.08$                                      \\
& gpt-4                       & 12.5\%                                & 6.9\%                                   & 0.8              & $0.37$                                      \\
\midrule

\textbf{Misinformation} & gpt-3.5-instruct-turbo      & \textbf{31.41\%}                               & 20.84\%                                 & {11.84}            & $5.80\times10^{-4}$           \\
& gpt-3.5-turbo (ChatGPT 3.5) & \textbf{28.99\%}                               & 15.5\%                                  & {20.54}            & $5.84\times10^{-6}$           \\
& gpt-4                       & \textbf{16.03\%  }                             & 4.2\%                                   & {30.14}            & $4.03\times10^{-8}$  \\

\bottomrule
\end{tabular}
\caption{\textbf{Error propagation by task.} False positive rate in downstream classification of counterspeech mentions, stratified by use-mention classification correctness. Statistics are reported separately by tasks.}
\label{tab:propagation_by_task}
\end{table*}

\begin{table*}[t]
\small
\centering
\begin{tabular}{p{0.27\columnwidth} | R{.32\columnwidth} | R{.25\columnwidth}  }
\toprule
\textbf{Target identity} & \textbf{False positive rate } & \textbf{Recall} \\
\midrule
Jewish       & 14.15\% & 98.11\%\\
People of color     &     9.09\% & 72.73\% \\
Muslims   &   6.80\% & 85.51\%\\
LGBT+     &   6.77\% & 80.45\%\\
Disabled  &   4.00\% & 96.00\%\\
Women   &     2.41\% & 85.71\%\\
Other   &    0.89\% & 56.25\% \\
Migrants  &   0.39\% & 53.64\%\\
\bottomrule
\end{tabular}
\caption{\textbf{Identity: False positive rate and recall.} False positive rate and recall in hate speech classification of counterspeech, stratified by target identity (gpt-3.5-turbo).}
\label{tab:recallbytarget}
\end{table*}

\begin{table*}[ht]
\small
    \begin{tabular}{p{0.25\linewidth}|>{\itshape}p{0.65\linewidth}}
    \textbf{Prompt} & {{\normalfont \textbf{Prompt text}}}\\
\toprule   
\textbf{Use-mention (hate speech)} & \noindent {The following text contains language that might be considered hateful. Classify the  text into one of the following two categories:}\\
& {A: use}\\
& {B: mention} \\

&  {Use means that the hateful language is actively used, and not referred to.}
{Examples of used:}\\
& {1. Using hateful language to express, incite, offend, or promote hate.} \\
& {2. Using hateful language to imply hate towards an identity group.}\\

&   {Mention means that hateful language is stated to illustrate properties of the hateful language itself.}
 {Examples of mentioned:} \\
& {1. Quoting, referring to, or paraphrasing hateful language to explain attitude towards it.}\\
& {2. Addressing hateful language when providing a counternarrative.}\\

&  {Output only A or B.}\\

&   {Text: <text>}\\

&  {Category: }\\
\midrule

\textbf{Use-mention (misinformation)} & The following text contains language that might be considered misinformation. Classify the text into one of the following two categories:\\
& A: use\\
& B: mention\\

&  Use means that misinformation is actively used, and not referred to.
Examples of used: \\ 
& 1. Using misinformation to deliberately deceive. \\
& 2. Using misinformative language to imply the truthfulness of misinformation. \\

& Mention means that misinformation is stated to illustrate properties of the misinformative language itself.
Examples of mentioned: \\
& 1. Quoting, referring to, or paraphrasing misinformation to express a lack of support for the misinformative language. \\
& 2. Addressing misinformation when providing a counternarrative.\\

&  Output only A or B.\\

&  Text: <text>\\

&   Category: \\

\midrule

\textbf{Downstream (hate speech)} & Classify the following text into one of the following two categories: [hateful, not hateful]. Hateful content expresses, incites, or promotes hate based on race, gender, ethnicity, religion, nationality, sexual orientation, disability status, or caste. \\

&  Text: <text> \\

& Category: \\

\midrule

\textbf{Downstream (misinformation)} &   Classify the following text into one of the following two categories: [misinformation, not misinformation]. Misinformation means that the text is either inaccurate or contains an element of truth but ignores critical facts that would give a different impression. \\

&  Text: <text> \\
&   Category: \\

\end{tabular}
\caption{\textbf{Prompt text (use-mention and downstream tasks).}}
\label{tab:prompts}
\end{table*}

\begin{table*}[ht]
\small
    \begin{tabular}{p{0.25\linewidth}|>{\itshape}p{0.65\linewidth}}
    \textbf{Prompt} & {{\normalfont \textbf{Prompt text}}}\\
\toprule   

\textbf{Few shot} &

Classify the following text into one of the following two categories: [hateful, not hateful]. Hateful content expresses, incites, or promotes hate based on race, gender, ethnicity, religion, nationality, sexual orientation, disability status, or caste. \\

&  Text: gay people are a burden to our society \\

& Category: hateful\\

& Text: I always feel unsafe when surrounded by Arabs\\

& Category: hateful\\

& Text: decent people are allowed here, so queers keep out\\

&  Category: hateful\\

&  Text: It is incredibly ignorant to claim that gay people are a burden to our society. This kind of discrimination has no place in our world.\\

& Category: not hateful\\

&  Text: Feeling unsafe just because someone is Arab is a form of prejudice and discrimination. I completely disapprove of that kind of mindset.\\

& Category: not hateful\\

& Text: <>\\
& Category: \\

\midrule

\textbf{Mitigation} &
Classify the following text into one of the following two categories: [hateful, not hateful]. Hateful content expresses, incites, or promotes hate based on race, gender, ethnicity, religion, nationality, sexual orientation, disability status, or caste. \\

&  Hateful is when hateful language is actively used, and not referred to. When language or arguments related to hate are just mentioned, the text is not hateful when debunking.\\

&  For example, text quoting, referring to, or paraphrasing hate speech to provide a counternarrative is not hateful.\\

&  Text: <text>\\

&  Category:\\

\midrule

\textbf{CoT~$+$~mitigation }  & Classify the following text into one of the following two categories: [hateful, not hateful]. Hateful content expresses, incites, or promotes hate based on race, gender, ethnicity, religion, nationality, sexual orientation, disability status, or caste. \\

 &  Use means that the hateful language is actively used, and not referred to. Mention means that hateful language is stated to illustrate properties of the hateful language itself.\\

&   Hateful is when hateful language is actively used, and not mentioned. When language or arguments related to hate are just mentioned, the text is not hateful when debunking.\\

& For example, text quoting, referring to, or paraphrasing hate speech to provide a counternarrative is not hateful. \\

& Text: gay people are a burden to our society\\

&   Answer: Let's think step by step.
Hateful language is actively used in order to express hate. So the answer is: hateful\\

&  Text: I always feel unsafe when surrounded by Arabs\\

&  Answer: Let's think step by step.
Hateful language is actively used in order to express hate. So the answer is: hateful\\

&  Text: decent people are allowed here, so queers keep out\\

&  Answer: Let's think step by step.
Hateful language is actively used in order to express hate. So the answer is: hateful\\

&  Text: It is incredibly ignorant to claim that gay people are a burden to our society. This kind of discrimination has no place in our world.\\

&   Answer: Let's think step by step.
Hateful language is mentioned. Mentioning is to provide a counternarrative. So the answer is: not hateful\\

&  Text: Feeling unsafe just because someone is Arab is a form of prejudice and discrimination. I completely disapprove of that kind of mindset.\\

& Answer: Let's think step by step.
Hateful language is mentioned. Mentioning is to provide a counternarrative. So the answer is: not hateful\\

& Text: <text>\\

& Answer: Let's think step by step.\\

\end{tabular}
\caption{\textbf{Prompt text (hate speech mitigations).}}
\label{tab:prompts_mitigations}
\end{table*}

\end{document}